\documentclass{article}
\usepackage{amsmath}
\usepackage{graphicx}

\usepackage{booktabs} 
\usepackage{multirow} 
\usepackage{xcolor}
\usepackage[preprint]{corl_2026} 

\usepackage{graphicx}
\usepackage{subfigure}
\usepackage{pifont}
\usepackage{setspace} 
\usepackage{colortbl}
\usepackage{dsfont}
\usepackage{bbm}
\usepackage{multirow}
\usepackage{amsmath}
\usepackage{amssymb}
\usepackage{mathtools}
\usepackage{amsthm}
\usepackage{booktabs}
\usepackage{caption}
\usepackage{wrapfig}

\RequirePackage{algorithm}
\RequirePackage{algorithmic}

\definecolor{light_green}{HTML}{E3F2D9}
\definecolor{deep_green}{HTML}{C8E5B3}
\definecolor{text_deep_green}{HTML}{72B73F}

\usepackage{hyperref}

\newcommand{\Checkmark}{\ding{51}}
\newcommand{\Crossmark}{\ding{55}}
\usepackage{pifont} 

\definecolor{mydarkblue}{rgb}{0,0.08,0.45}
\definecolor{darkred}{rgb}{0.55,0,0}

\hypersetup{
    colorlinks=true,
    citecolor=mydarkblue, 
    linkcolor=darkred    
}

\title{GeoSem-WAM:
Geometry- and Semantic-Aware World Action Models}

\author{%
  Fulong Ma\textsuperscript{1}\thanks{Equal contribution.}\quad Daojie Peng\textsuperscript{1}\footnotemark[1]\quad Wenjun Yue\textsuperscript{4}\quad Jiahang Cao\textsuperscript{2}\\
    \textbf{Bintao Wang\textsuperscript{5}\quad Qiang Zhang\textsuperscript{1,3,6}\quad Jun Ma\textsuperscript{1}\thanks{Corresponding author: \texttt{jun.ma@ust.hk}}} \\
\textsuperscript{1}HKUST(GZ)\quad \textsuperscript{2}HKU\quad \textsuperscript{3}USTC\quad \textsuperscript{4}OC
\quad \textsuperscript{5}SDU\quad \textsuperscript{6}X-Humaniod
}

\begin{document}
\maketitle

\vspace{-20pt}
\begin{abstract}
Recent World Action Models (WAMs) have demonstrated impressive capabilities in embodied decision-making. However, whether their effectiveness stems from explicit future imagination during inference or representation learning induced by predictive training remains an open question. Emerging evidence suggests the primary advantage lies in learning robust latent representations rather than generating future observations at test time. Nevertheless, existing WAMs mainly rely on RGB-based future prediction, which provides limited structural and spatial understanding of complex environments. To address this, we propose a structured world modeling framework that enhances latent representations through geometric and semantic supervision. Alongside future RGB prediction, our model introduces two auxiliary prediction branches for future geometry and semantic representations, enabling it to jointly capture scene dynamics, spatial geometry, and semantic context within a unified latent space. Crucially, our approach preserves efficient inference by avoiding explicit future rollout or video generation at test time. Extensive experiments show that incorporating structured world supervision consistently improves action prediction accuracy, scene understanding, and robustness under challenging embodied scenarios, highlighting its potential for advancing scalable and efficient WAMs.

\end{abstract}


\section{Introduction}
World Action Models (WAMs) have emerged as a transformative paradigm for embodied intelligence, enabling agents to learn predictive representations of environmental dynamics from large-scale interaction data—distinct from conventional policy learning that directly maps observations to actions \cite{brohan2022rt,zitkovich2023rt,kim2024openvla,bu2025univla,peng2026structured,intelligence2025pi}, WAMs leverage predictive world modeling as an auxiliary objective to enhance decision-making through future-aware representation learning. This approach has achieved remarkable performance across diverse embodied tasks, underscoring the critical role of predictive world modeling in learning robust action policies \cite{yuan2026fast,cen2025worldvla,peng2026attena+,bi2025motus, hu2026navthinker}, particularly with the advancement of vision-language-action (VLA) models that have established foundational robotic policies (e.g., RT-1 \cite{brohan2022rt}, RT-2 \cite{zitkovich2023rt}, OpenVLA \cite{kim2024openvla}) and generalist frameworks (e.g., Octo \cite{team2024octo}, $\pi_0$ \cite{black2024pi_0}) to unify perception, language, and action, with subsequent works optimizing fine-tuning and action tokenization for real-world deployment \cite{kim2025fine,pertsch2025fast}. 

Yet despite this progress, a fundamental question remains understudied: \textit{why do World Action Models work?} Early WAM designs assumed that explicit future imagination during inference \cite{feng2025vidar,zhu2025unified}, generating future trajectories or visual observations \cite{ye2026world,li2026causal} to plan ahead drove their success, but recent evidence increasingly points to a different core benefit: the dynamics-aware latent representations learned through predictive supervision during training, rather than test-time future imagination \cite{yuan2026fast}. In essence, future prediction acts as a structured self-supervised objective. Existing WAMs predominantly rely on RGB-based future prediction, which provides only limited structural understanding of complex environments. Appearance-based supervision lacks explicit geometric reasoning and high-level semantic awareness that are critical for embodied agents operating in real-world scenarios \cite{huang2025spatial}, resulting in latent representations that capture short-term visual dynamics but fail to encode richer environmental structure and object-level semantics.

To address this gap, we propose a structured world modeling framework that augments WAMs with geometry-aware and semantic-aware predictive supervision. Beyond standard future RGB prediction, our framework introduces two auxiliary branches: a geometry prediction branch to learn spatial geometry and 3D structural consistency, and a semantic prediction branch to capture object-level semantics and scene context \cite{peng2026structured}. By jointly modeling future appearance, geometry, and semantics, our framework learns a more structured latent world representation that better captures the underlying properties of embodied environments—all while preserving the efficient inference paradigm of modern WAMs. Unlike methods requiring computationally expensive future rollout or iterative generation at test time, our approach uses structured predictive supervision only during training, directly predicting actions at inference to balance performance and latency, aligning with ongoing efforts to accelerate VLA inference via token pruning, cache optimization, and dynamic compression \cite{xu2026vla,tan2025think,li2025sp}.
Extensive evaluations across diverse embodied interaction tasks confirm that our approach consistently improves action prediction accuracy, robustness, and scene understanding, particularly in challenging scenarios involving occlusions, object interactions, and complex environmental dynamics. Our contributions are summarized as follows:
\begin{enumerate}
    \item We revisit the core value of predictive world modeling in WAMs, providing a clear perspective that its primary benefit stems from training-phase representation learning (i.e., inducing dynamics-aware latent features via predictive supervision) rather than explicit future imagination or rollout during test time, which clarifies the underpinning mechanism of WAM effectiveness and guides more efficient model design.
    \item We propose a novel structured world modeling framework that enriches WAM supervision with multi-modal predictive signals, integrating future RGB, geometry, and semantic prediction into a unified framework. This design explicitly encourages the model to learn spatial geometry, 3D structural consistency, and object-level semantics, addressing the limitation of RGB-only supervision in capturing complex environmental structure.
    \item We demonstrate through comprehensive simulation and real-world experiments that our structured supervision strategy consistently enhances embodied decision-making performance across diverse tasks, while maintaining efficient test-time inference by avoiding explicit future generation. This balance of performance and efficiency makes our framework practical for real-world robotic deployment, with additional analyses verifying the complementary value of geometric and semantic supervision.
\end{enumerate}

\begin{figure}[htbp]      
    \centering           
    \scalebox{1.0}[0.95]{\includegraphics[width=1.0\textwidth]{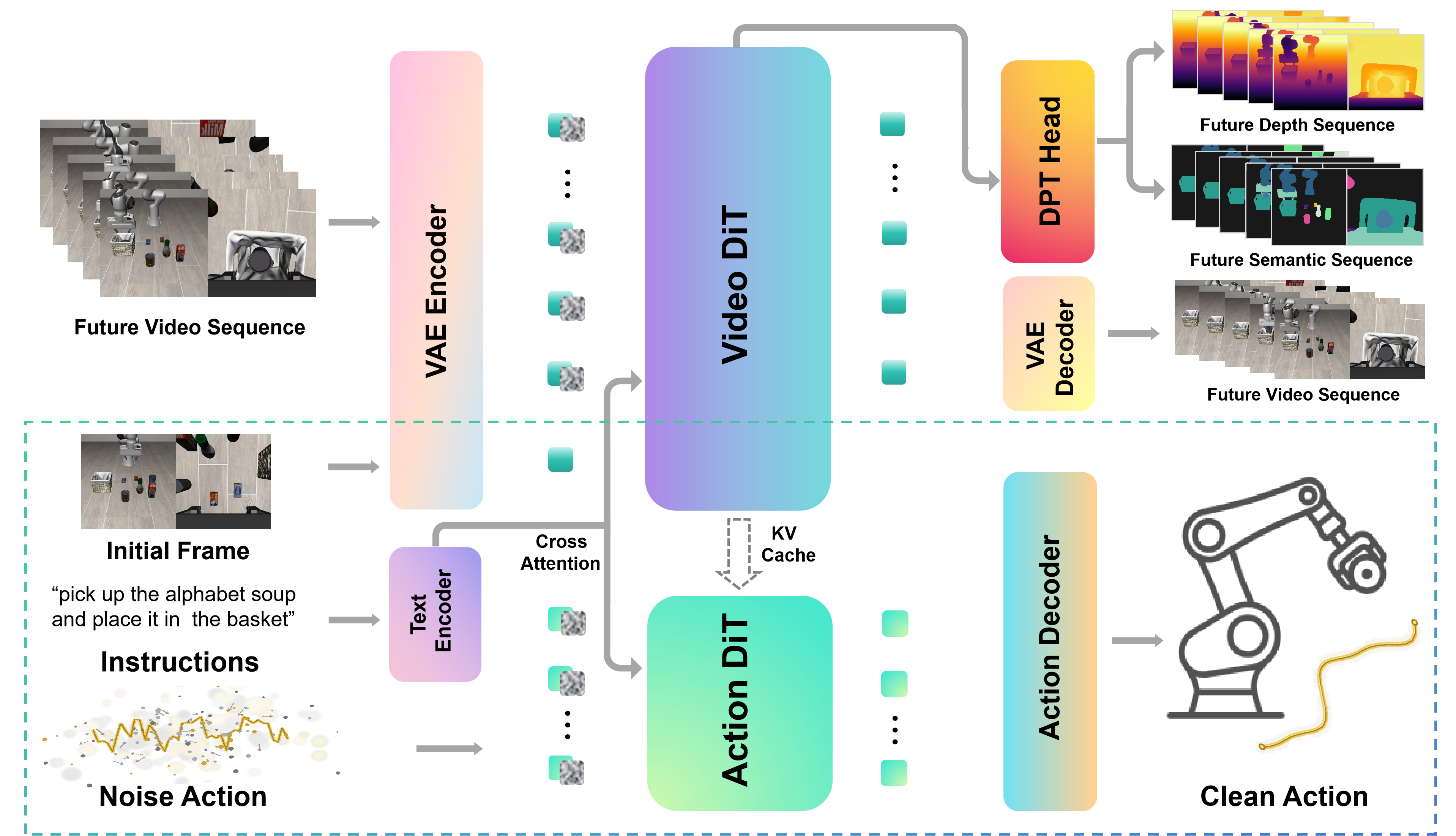}}  
    \caption{\footnotesize Overview of the architecture of our method. The overall figure represents the training phase, and the part within the dashed box represents the model inference stage.}
    \label{fig:architecture} 
\end{figure}

\vspace{-10pt}
\section{Related Works}

\vspace{-5pt}
\subsection{Vision-Language-Action (VLA) Policies and World Action Models (WAMs)}
Vision-Language-Action (VLA) models serve as the cornerstone of modern embodied intelligence, unifying visual perception, natural language grounding, and robotic action generation. Representative works such as RT-1 \cite{brohan2022rt}, RT-2 \cite{zitkovich2023rt}, and OpenVLA \cite{kim2024openvla} successfully transfer web-scale knowledge to real-world robotic control, while generalist frameworks like Octo \cite{team2024octo} further expand multi-task adaptability. World Action Models (WAMs) integrate world modeling with action generation, leveraging predictive supervision to learn environmental dynamics and improve decision-making \cite{cen2025worldvla,bi2025motus}. Most existing WAMs rely on RGB-based future prediction as their core supervision, but recent studies confirm that their key advantage lies in training-phase dynamics-aware representation learning rather than test-time explicit future imagination \cite{yuan2026fast}. Unlike prior works focusing on VLA deployment optimization (e.g., inference acceleration \cite{xu2026vla} or fine-tuning \cite{kim2025fine}) or RGB-only WAM designs, our work enhances training supervision with geometric and semantic cues to learn more structured latent representations.

\subsection{Imitation Learning and Robotic Data Learning}
Imitation learning is the core technical support for robotic policy training from demonstration data. Early researches focus on heterogeneous demonstration screening, state adaptive weighting and coarse-to-fine learning strategies to improve imitation efficiency \cite{mandlekar2019iris}. Meanwhile, large-scale robotic datasets including BridgeData V2 \cite{walke2023bridgedata} and standardized evaluation benchmarks such as CALVIN \cite{mees2022calvin}, LIBERO \cite{liu2023libero} provide unified training and verification platforms for embodied policy. In addition, data quality enhancement and automatic data curation methods also greatly facilitate scalable robotic model training \cite{hejna2025robot,chen2025robotwin}. Our structured world modeling can serve as an effective representation enhancement module, which can be seamlessly embedded into imitation learning pipelines to excavate deeper structural information from limited demonstration data.

\subsection{Scaling Laws and Foundation Model Representation Learning}
Scaling law research in natural language processing reveals that model capability can be steadily promoted through reasonable allocation of parameters, data and computing resources \cite{kaplan2020scaling,hoffmann2022training,gordon2021scaling}. Such conclusions also provide important guidance for the development of robotic foundation models. On the basis of large-scale pre-trained visual-language models such as CLIP \cite{radford2021learning} and EVA-CLIP \cite{EVA_CLIP}, embodied models gradually migrate general visual-text alignment knowledge to physical interaction scenarios. Our work conforms to this development trend, and enhances the task-specific structured representation ability of robotic foundation models through customized multi-modal world prediction supervision, without blindly expanding model scale and training data volume.


\vspace{-5pt}
\section{Methodology}
\label{sec:methodology}



\vspace{-5pt}
\subsection{Overview}
\label{subsec:overview}


GeoSem-WAM is motivated by the promise of world modeling for learning richer downstream representations. Beyond standard future pixel prediction, we introduce auxiliary geometry and semantic segmentation branches during training. Similar to Fast-WAM \cite{yuan2026fast}, GeoSem-WAM jointly learns video generation, action prediction, and geometric-semantic understanding, forcing the backbone network to capture physically grounded motion and spatial-semantic layouts. During inference, GeoSem-WAM avoids explicit future sequence prediction. Instead, it processes only the first observation’s latent tokens in a single forward pass to directly generate actions, eliminating the computational overhead of future rollouts. The DPT auxiliary branches are also discarded at deployment. Importantly, neither geometry nor semantic annotations are used as model inputs. This design mirrors human cognition: relying solely on raw visual observation while internally reasoning about geometry and semantics to achieve superior task performance.

\subsection{Architecture}
\label{subsec:architecture}

GeoSem-WAM is constructed upon the video Diffusion Transformer (DiT) of Wan2.2-5B \cite{wan2025wan}, which acts as the world modeling backbone. The pretrained text encoder and video VAE from the same model are also reused: task instructions are encoded using the native T5 encoder and delivered to all tokens via cross-attention, whereas visual observations are transformed into latent video tokens through the pretrained VAE. Built on this backbone, we introduce an action expert DiT, similar in architecture but differing in size, designed for action chunk generation. Furthermore, we incorporate DPT-style \cite{ranftl2021vision} geometry prediction and semantic segmentation branches. 
The overall model adopts a Mixture-of-Transformer (MoT) architecture with shared attention between the video and action branches, as shown in Fig. \ref{fig:architecture}. The dashed box denotes the model's inputs and network architecture at inference stage.

\textbf{Latent World Modeling.}
We model future video dynamics in the latent space of a pretrained VAE.
Let \(z^{\mathrm{gt}}_{t:t+K}\) denote the ground-truth future video latents.
During training, we sample a noise level \(\sigma \in [0,1]\) and corrupt the target video latents with Gaussian noise \(\epsilon_z \sim \mathcal{N}(0,I)\):
\begin{equation}
   z^\sigma_{t:t+K}
=
(1-\sigma)z^{\mathrm{gt}}_{t:t+K}
+
\sigma \epsilon_z .
\end{equation}

Conditioned on the current observation and language instruction, the video DiT predicts the flow target:
\begin{equation}
    \hat{v}_z
=
f^{\mathrm{rgb}}_{\theta}
\left(
z^\sigma_{t:t+K}, \sigma, c
\right),
\quad
v_z
=
\epsilon_z
-
z^{\mathrm{gt}}_{t:t+K},
\end{equation}

where \(c\) denotes the conditioning context, including the current visual observation and language instruction.
The video modeling objective is:
\begin{equation}
    \mathcal{L}_{\mathrm{rgb}}
=
\left\|
\hat{v}_z - v_z
\right\|_2^2 .
\end{equation}

\textbf{Action Modeling.}
The action branch predicts a future action chunk through denoising. During training, we corrupt the ground-truth action sequence \(a^{\mathrm{gt}}_{t:t+H-1}\) with Gaussian noise \(\epsilon_a\):
\begin{equation}
a^\sigma_{t:t+H-1}
=
(1-\sigma)a^{\mathrm{gt}}_{t:t+H-1}
+
\sigma \epsilon_a,
\quad
\epsilon_a \sim \mathcal{N}(0,I).
\end{equation}

Conditioned on the latent world representation \(z_t\), the action DiT predicts the flow target:
\begin{equation}
\hat{v}_a
=
f^{\mathrm{act}}_{\phi}
\left(
a^\sigma_{t:t+H-1}, \sigma, z_t
\right),
\quad
v_a
=
\epsilon_a
-
a^{\mathrm{gt}}_{t:t+H-1}.
\end{equation}

The action objective is:
\begin{equation}
    \mathcal{L}_{\mathrm{act}}
=
\left\|
\hat{v}_a - v_a
\right\|_2^2 .
\end{equation}

At inference time, the action chunk is initialized from Gaussian noise and iteratively denoised conditioned on \(z_t\), without explicitly generating future video frames.

\begin{wrapfigure}{r}{0.5\textwidth}  
    \centering
    \scalebox{1.0}[0.9]{
        \includegraphics[width=0.5\textwidth]{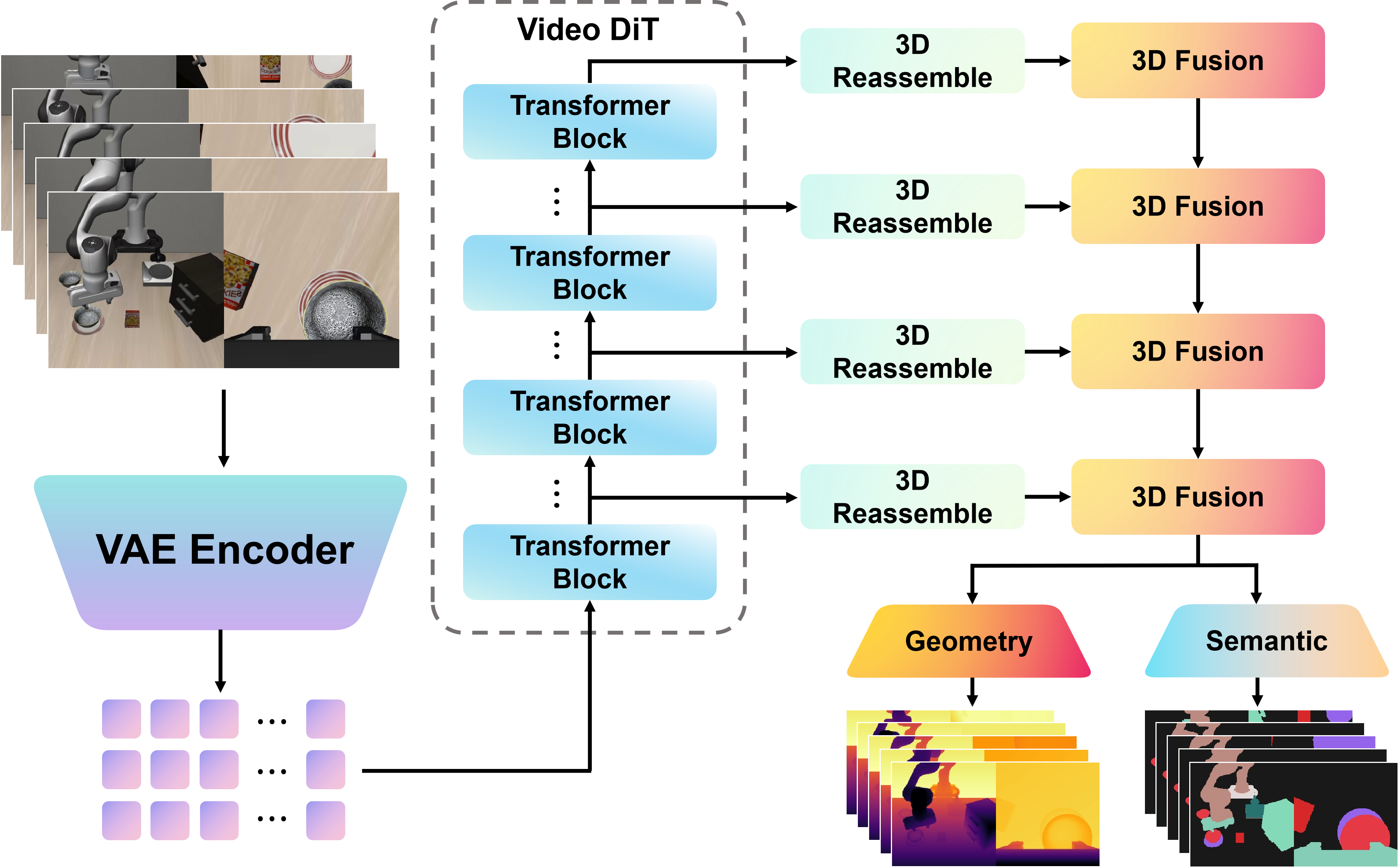}
    }
    \caption{\footnotesize The architecture of DPT auxiliary head.}
    \label{fig:dpt}
\end{wrapfigure}

\paragraph{Dense Structured World Supervision.}
To encourage the learned world representation to encode both geometric structure and object-level semantics, we introduce dense auxiliary supervision on the video latent tokens. We implement the auxiliary branch with a DPT-style \cite{ranftl2021vision} dense prediction head. This DPT-style head aggregates intermediate video tokens from multiple Transformer blocks. These multi-level features are projected, fused, and decoded into dense spatial predictions, allowing the auxiliary supervision to leverage both low-level spatial details and high-level semantic abstractions. The details of this DPT-style dense prediction branch are illustrated in Fig. \ref{fig:dpt}. Specifically, the input video is first encoded into tokens via a VAE encoder. After these tokens are processed through multiple Transformer stages, the architecture reassembles the multi-stage tokens into multi-resolution, image-like representations. These representations are then progressively fused and upsampled through fusion modules, and finally decoded by the geometry and semantic heads to yield fine-grained predictions. To better accommodate video inputs, we extend the original reassemble and fusion modules to a 3D reassemble module and a 3D fusion module.

Let \(z_{\tau}\) denote the latent representation at a future prediction step \(\tau \in \{t+1,\ldots,t+K\}\).
For geometry supervision, we attach a geometry prediction head \(H_{\mathrm{geo}}\) to estimate the future geometry information $\hat{o}^{\mathrm{geo}}_{\tau}$,
and the geometry branch is trained with an \(L_1\) reconstruction objective:
\begin{equation}
\mathcal{L}_{geo}
= \frac{1}{K}\sum\nolimits_{\tau}
\|\hat{o}^{geo}_\tau - o^{geo}_\tau\|_1.
\end{equation}
For semantic supervision, we attach a semantic prediction head \(H_{\mathrm{sem}}\) to predict dense semantic logits $\hat{o}^{\mathrm{sem}}_{\tau}$,
the semantic branch is optimized using pixel-wise cross-entropy:
\begin{equation}
\mathcal{L}_{sem}
= \frac{1}{K}\sum\nolimits_{\tau}
\mathrm{CE}(\hat{o}^{sem}_\tau, o^{sem}_\tau).
\end{equation}

\textbf{Unified Training Objective.} The overall training objective jointly optimizes RGB prediction, geometry prediction, semantic prediction, and action prediction:

\begin{equation}
\mathcal{L}
=
\lambda_{rgb}\mathcal{L}_{rgb}
+
\lambda_{geo}\mathcal{L}_{geo}
+
\lambda_{sem}\mathcal{L}_{sem}
+
\lambda_{act}\mathcal{L}_{act}
\end{equation}

where
$\lambda_{rgb}$,
$\lambda_{geo}$,
$\lambda_{sem}$,
and
$\lambda_{act}$
denote balancing coefficients for different objectives.

\vspace{-5pt}
\section{Experiments}
\label{sec:experiments}

\vspace{-3pt}
\subsection{Experimental Setup}
\textbf{Simulation Environment.} We conduct experiments on two commonly adopted simulation benchmarks,
LIBERO~\citep{liu2023libero} and RoboTwin~\citep{chen2025robotwin}.
LIBERO includes four task suites, namely LIBERO-Spatial, LIBERO-Object,
LIBERO-Goal, and LIBERO-Long. Each suite provides 500 expert demonstrations
covering 10 tasks, enabling evaluation of policy generalization across spatial
configurations, object categories, goal specifications, and long-horizon
execution. RoboTwin is a real-to-sim benchmark designed for bimanual robotic
manipulation. It provides an easy setting with in-domain layouts and a more
challenging setting with domain randomization, where variations are introduced
through scene clutter, background textures, illumination, and tabletop height.
We evaluate our approach on a diverse set of tasks and use success rate (SR) as
the evaluation metric for both benchmarks.


\begin{figure}[htbp]      
    \centering           
    \scalebox{0.95}[0.90]{
    \includegraphics[width=1.0\textwidth]{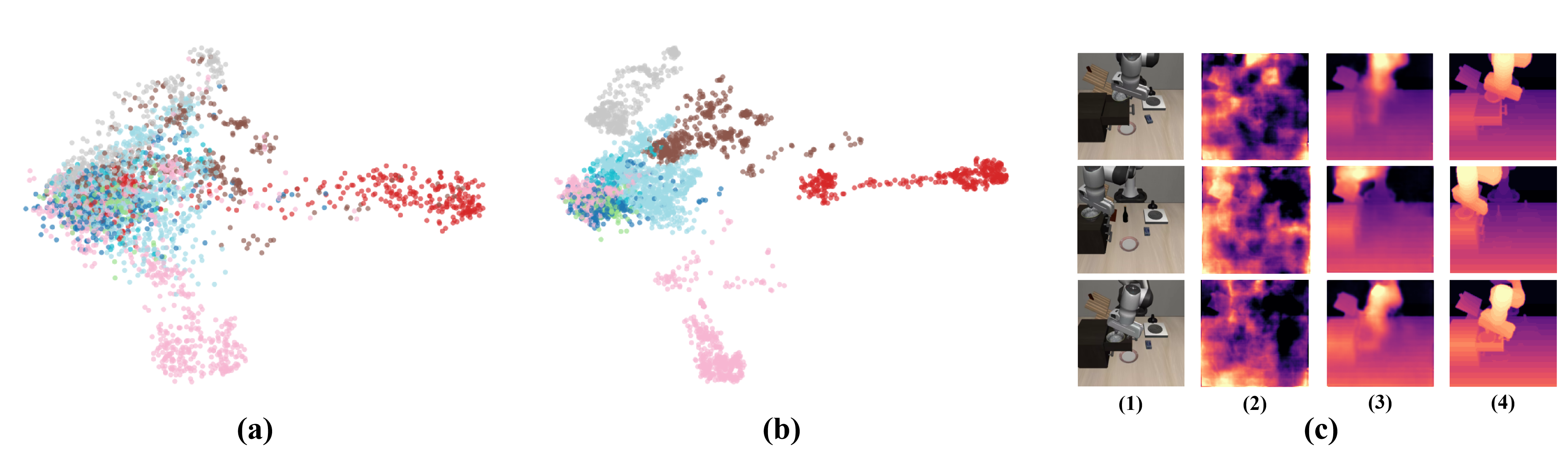}}
    \caption{\footnotesize Fig. (a) and (b): Middle layer Video DiT token embeddings colored by semantic class. GeoSem-WAM yields clearer semantic clustering than baseline. Fig. (c): Frozen-backbone depth probing on LIBERO. GeoSem-WAM yields more accurate depth predictions from Video DiT tokens, suggesting richer geometry-aware latent representations.}
    \label{fig:dep_sem_compare} 
\end{figure}

\vspace{-5pt}

\subsection{Comparisons with State-of-the-Art Methods}

\textcolor{black}{\textbf{LIBERO.} 
Each task is evaluated for 50 trials under different random seeds, and we report the success rate of each task suite as well as the mean success rate across the four suites. As shown in Table \ref{tab:libero_results}, our GeoSem-WAM achieves an overall average success rate of 98.55\%, demonstrating strong performance across all task categories. Compared with the baseline method Fast-WAM, the average success rate improves from 97.60\% to 98.55\%, validating the effectiveness of introducing explicit geometry and semantic supervision for future video prediction.}

\begin{table*}[]
    \centering
    \caption{Comparisons with SOTA methods on LIBERO benchmark.}
    \label{tab:libero_results}
    \renewcommand{\arraystretch}{0.82} 
    \setlength{\aboverulesep}{1.5pt}    
    \setlength{\belowrulesep}{1.5pt}    
    \resizebox{1.0\textwidth}{!}{
    \begin{tabular}{l|c|ccccc}
    \toprule
        \textbf{Method} & \textbf{Paradigm} & \textbf{Spatial SR (\%)} 
        \textcolor{teal}{$\uparrow$}& \textbf{Object SR (\%)} 
        \textcolor{teal}{$\uparrow$}& \textbf{Goal SR (\%)} 
        \textcolor{teal}{$\uparrow$}& \textbf{Long SR (\%)} 
        \textcolor{teal}{$\uparrow$}& \textbf{Average SR (\%)} \textcolor{teal}{$\uparrow$} \\
        \midrule
        OpenVLA \cite{kim2024openvla} & VLA & 84.7 & 88.4 & 79.2 & 53.7 & 76.50 \\
        VLA-Cache \cite{xu2026vla} & VLA & 83.8 & 85.8 & 76.4 & 52.8 & 74.70 \\
        FlashVLA \cite{tan2025think} & VLA & 84.2 & 86.4 & 75.4 & 51.4 & 74.35 \\
        SP-VLA \cite{li2025sp} & VLA & 75.4 & 85.6 & 84.4 & 54.2 & 74.90 \\
        WorldVLA \cite{cen2025worldvla} & VLA & 85.6 & 89.0 & 82.6 & 59.0 & 79.05 \\
        NORA-Long \cite{hung2025nora} & VLA & 92.2 & 95.4 & 89.4 & 74.6 & 87.90 \\
        SmolVLA \cite{shukor2025smolvla} & VLA & 93.0 & 94.0 & 91.0 & 77.0 & 88.75 \\
        CogACT \cite{li2024cogact} & VLA & 97.2 & 98.0 & 90.2 & 88.8 & 93.55 \\
        $\pi_0$ + FAST \cite{pertsch2025fast} & VLA & 96.4 & 96.8 & 88.6 & 60.2 & 85.50 \\
        $\pi_0$ \cite{black2024pi_0} & VLA & 96.8 & 98.8 & 95.8 & 85.2 & 94.15 \\
        $\pi_{0.5}$ \cite{intelligence2025pi} & VLA & 98.8 & 98.2 & \underline{98.0} & 92.4 & 96.85 \\
        UniVLA \cite{bu2025univla} & VLA & 96.5 & 96.8 & 95.6 & 92.0 & 95.23 \\
        VLA-ADP \cite{pei2025action} & VLA & \textbf{99.0} & 98.2 & 96.8 & 91.2 & 96.30 \\
        OpenVLA-OFT \cite{kim2025fine} & VLA & 97.6 & 98.4 & 97.9 & 94.5 & 97.10 \\
        Motus \cite{bi2025motus} & WAM & 96.8 & \underline{99.8} & 96.6 & \underline{97.6} & 97.70 \\
        LingBot-VA \cite{li2026causal} & WAM & \underline{98.5} & 99.6 & 97.2 & \textbf{98.5} & \underline{98.50} \\
        Fast-WAM \cite{yuan2026fast} & WAM & 97.2 & \textbf{100.0} & 97.0 & 95.2 & 97.60 \\
        \midrule
        \textbf{GeoSem-WAM (ours)} & WAM & \textbf{99.0} & \textbf{100.0} & \textbf{98.2} & 97.0 & \textbf{98.55} \\
        \bottomrule
    \end{tabular}}
\end{table*}

\textcolor{black}{\textbf{RoboTwin 2.0} 
On the RoboTwin 2.0 dataset, we evaluate 50 tasks under both the clean and random settings. Table \ref{tab:robotwin} reports the success rates under the clean and random settings, as well as the overall average success rate. As shown in Table 2, our GeoSem-WAM achieves a new state-of-the-art average success rate of 92.52\%. Compared with the base model Fast-WAM, it improves the average success rate by 0.8\% and outperforms the previous best method, LingBot-VA, without requiring any embodied pre-training. For the specific success rates of each task, please refer to Table \ref{tab:robo_twin_sim}.}

\begin{table}[t]
\centering
\caption{Performance on RoboTwin 2.0 Compared with SOTA Methods.}
\label{tab:robotwin} 
\renewcommand{\arraystretch}{0.85} 
\setlength{\aboverulesep}{1.5pt}    
\setlength{\belowrulesep}{1.5pt}    
\resizebox{1.0\textwidth}{!}{
\begin{tabular}{l|c|c|ccc}
    \toprule
    Method & Paradigm & Embodied PT. & \textbf{Clean SR (\%)} 
    \textcolor{teal}{$\uparrow$}& \textbf{Random SR (\%)} 
    \textcolor{teal}{$\uparrow$}& \textbf{Average SR (\%) \textcolor{teal}{$\uparrow$}} \\
    \midrule
    $\pi_0$ \cite{black2024pi_0} & VLA & \Checkmark & 65.92 & 58.40 & 62.20 \\
    $\pi_{0.5}$ \cite{intelligence2025pi} & VLA & \Checkmark & 82.74 & 76.76 & 79.75 \\
    X-VLA \cite{zheng2025x} & VLA & \Checkmark & 72.90 & 72.80 & 72.85 \\
    UWM \cite{zhu2025unified} & WAM & \Checkmark & 81.70 & 78.60 & 80.15 \\
    GigaWorld-Policy \cite{ye2026gigaworld} & WAM & \Checkmark & 87.00 & 85.00 & 86.00 \\
    Motus \cite{bi2025motus} & WAM & \Checkmark & 88.66 & 87.02 & 87.80 \\
    X-WAM \cite{guo2026unified} & WAM & \Checkmark & 89.80 & 90.70 & 90.25 \\
    LingBot-VA \cite{li2026causal} & WAM & \Checkmark & \underline{92.90} & 91.50 & \underline{92.20} \\
    Fast-WAM \cite{yuan2026fast} & WAM & \Crossmark & 91.88 & \underline{91.78} & 91.80 \\
    \midrule
    \textbf{GeoSem-WAM (ours)} & WAM & \Crossmark & \textbf{92.94} & \textbf{92.14} & \textbf{92.52} \\
    \bottomrule
\end{tabular}}
\end{table}


Furthermore, we analyze GeoSem-WAM from both semantic and geometric perspectives, with qualitative visualizations shown in Fig. \ref{fig:dep_sem_compare}. In the semantic experiment, as illustrated in Figures 3(a) and 3(b), token visualizations from the intermediate layers of ViT show that GeoSem-WAM exhibits clearer class clustering compared to Fast-WAM, indicating that our method achieves better semantic understanding. Additionally, we freeze the backbone and train a simple depth probe using only the intermediate tokens. The results are shown in Fig. 3(c), where columns 1 to 4 represent the RGB image, the depth map predicted by the depth probe based on Fast-WAM, the depth map predicted by the depth probe based on GeoSem-WAM, and the ground truth depth map, respectively. From Fig. 3(c), it can be observed that the latent tokens of GeoSem-WAM predict depth maps closer to the GT depth maps, whereas those of the baseline method produce blurry depth maps. This demonstrates that under the constraints of both geometric and semantic branches, the model's latent representation is enhanced for both semantic and geometric understanding. Real robot experiments further validate that our method achieves superior performance on tasks involving semantic and geometric changes, as detailed in Section \ref{experiment:real world}.

\subsection{Ablation Study}
\label{subsec:ablation_study}


We conduct ablation studies on the LIBERO benchmark to validate our auxiliary branches, using Fast-WAM as the baseline (Table \ref{tab:ablation_study}). Introducing only geometry supervision improves the average success rate from 97.6\% to 98.2\% (+0.61\%), while only semantic supervision yields 98.1\% (+0.51\%). Combining both branches achieves the most significant improvement, raising the success rate to 98.6\% (+1.02\%).
These results indicate that explicit geometric and semantic supervision for future video prediction both contribute positively, with their combination yielding the best performance. This aligns with intuition: for robotic manipulation, geometry and semantics correspond to spatial motion perception during execution and task-level logical reasoning, respectively. Together, they form the foundation for accurate and appropriate grasping, thereby enhancing the model’s spatial perception and reasoning capabilities. 

\begin{table}[t]
    \centering
    \caption{\footnotesize Component analysis of different structured world supervision objectives on the LIBERO benchmark.}
    \label{tab:ablation_study}
    \renewcommand{\arraystretch}{0.85} 
    \setlength{\aboverulesep}{1.5pt}    
    \setlength{\belowrulesep}{1.5pt}    
    \resizebox{0.9\textwidth}{!}{
    \begin{tabular}{lccccc}
        \toprule
        Method & RGB & Geometry & Semantic &Average SR (\%) \textcolor{teal}{$\uparrow$} & $\Delta$SR (\%) \textcolor{teal}{$\uparrow$}\\
        \midrule
        RGB-only WAM & $\checkmark$ & $\times$ & $\times$ & 97.6 & - \\
        + Geometry-aware & $\checkmark$ & $\checkmark$ & $\times$ & 98.2 &\textcolor{teal}{+0.61} \\
        + Semantic-aware & $\checkmark$ & $\times$ & $\checkmark$ & 98.1 & \textcolor{teal}{+0.51} \\
        + Geometry- and Semantic-Aware & $\checkmark$ & $\checkmark$ & $\checkmark$ & \textbf{98.6} & \textcolor{teal}{+1.02}\\
        \bottomrule
    \end{tabular}
    }
\end{table}

\vspace{-5pt}
\subsection{Real-World Experiments on Franka Emika Panda Robot}
\label{experiment:real world}
\begin{figure}[t]
    \centering
    \includegraphics[width=1.0\linewidth]{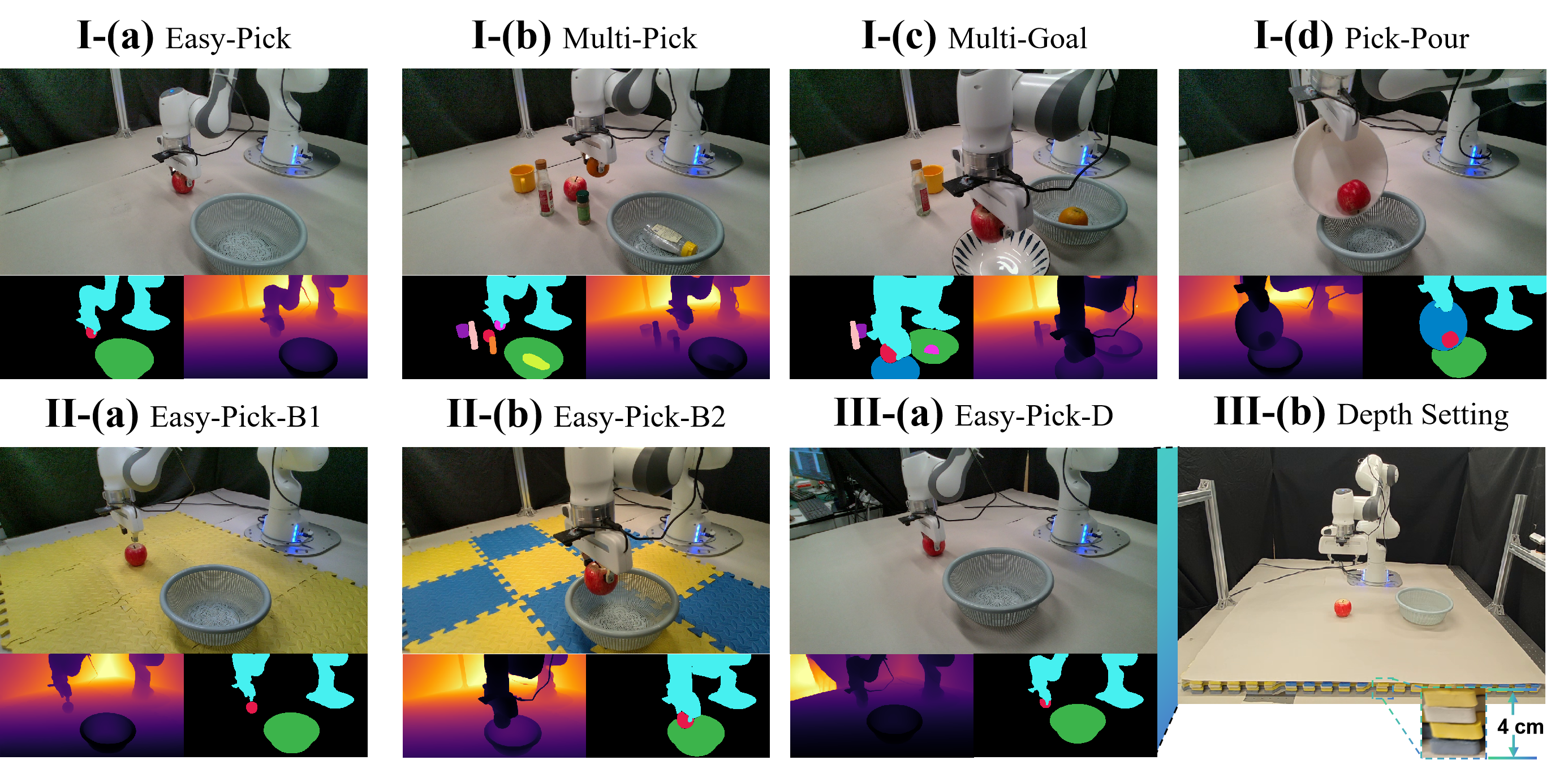}
    \caption{\footnotesize \textbf{Real-world manipulation experiments overview.} 
    (I) Four core tasks: Easy-Pick, Multi-Pick, Multi-Goal, and Pick-Pour, each shown with RGB, depth, and semantic observations. 
    (II) Background generalization tests (Easy-Pick-B1/B2) on different mat patterns. 
    (III) Height generalization tests: standard setup (Easy-Pick-D) vs. 4 cm elevated platform.}
    \label{fig:real_franka}
\end{figure}

\vspace{-5pt}
We conduct real robotic validation on a Franka Emika Panda platform with four manipulation tasks of escalating difficulty, as depicted in Figure~\ref{fig:real_franka}-I: 
(a) Easy-Pick for single-object pick-and-place, 
(b) Multi-Pick handling objects amid multiple distractors, 
(c) Multi-Goal multi-object placement to different target containers, 
(d) Pick-Pour cross-bowl apple pouring, which involves long-horizon pick-place-pour coordination. 
We collect 50 human teleoperation trajectories per task. Raw sequences are preprocessed by discarding idle frames and smoothing action sequences to facilitate stable and efficient model training. We fine-tune GeoSem-WAM following the Fast-WAM training paradigm utilizing two NVIDIA H800 GPUs. During inference, the model is deployed on an RTX 4090 GPU. Each task undergoes 50 repeated trials to calculate average success rate Average SR.

To further evaluate the robustness of our policy, we design two additional generalization test settings, as shown in Figure~\ref{fig:real_franka}-II and \ref{fig:real_franka}-III. 
Figure~\ref{fig:real_franka}-II presents the background generalization tests on the Easy-Pick task, with two different mat backgrounds (uniform yellow and patterned blue-yellow) to assess how visual distractions affect performance. 
Figure~\ref{fig:real_franka}-III illustrates the height generalization tests, where we compare the standard setup (Easy-Pick-D) with an elevated platform setup featuring a 4 cm height difference, verifying the policy's adaptability to geometric variations in the workspace.

Quantitative results across all real-world scenarios are summarized in Table~\ref{tab:real_experiments_results}. The proposed GeoSem-WAM consistently outperforms the Fast-WAM baseline across all tested tasks and generalization settings. Overall, the average success rate improves from 88.9\% to 95.4\%, representing a clear performance gain of +6.6\%. 
Notably, the largest improvements are observed in the generalization and challenging multi-step tasks:
(1) In the \textbf{background and height generalization tests}, GeoSem-WAM achieves gains of +10\%, +8\%, and +12\% on Easy-Pick-B1, Easy-Pick-B2, and Easy-Pick-D, respectively, effectively mitigating performance drops caused by visual distractions and geometric variations. 
(2) For the \textbf{multi-object and long-horizon tasks}, it yields consistent improvements of +6\% on Multi-Pick, +6\% on Multi-Goal, and +4\% on Pick-Pour, demonstrating stronger robustness in scenarios requiring complex spatial reasoning and sequential action planning. 
Across all setups, the single-object Easy-Pick task already reaches perfect performance (100\%) with both methods, confirming that our geometric-semantic priors do not degrade basic manipulation capabilities. These results validate that incorporating fused geometric and semantic features significantly enhances the policy's reliability, adaptability, and generalization in practical robotic manipulation scenarios.

\begin{table}[t]
    \centering
    \caption{Quantitative results of real Franka robot experiments. All tasks are evaluated over 50 independent trials.}
    \label{tab:real_experiments_results} 
    \renewcommand{\arraystretch}{0.85} 
    \setlength{\aboverulesep}{1.5pt}    
    \setlength{\belowrulesep}{1.5pt}    
    \resizebox{1.0\textwidth}{!}{
    \begin{tabular}{c|cccc|ccc|c}
        \toprule
        Model & Easy-Pick & Easy-Pick-B1 & Easy-Pick-B2 & Easy-Pick-D & Multi-Pick & Multi-Goal & Pick-Pour & Average SR  \\
        \midrule
        Fast-WAM  & 100 & 86 & 86 & 80 & 92 & 90 & 88 & 88.9 \\
        GeoSem-WAM  & 100 & 96 & 94 & 92 & 98 & 96 & 92 & 95.4 \\
        \midrule
        Improvement  & \textcolor{teal}{0} & \textcolor{teal}{+10} & \textcolor{teal}{+8} & \textcolor{teal}{+12} & \textcolor{teal}{+6} & \textcolor{teal}{+6} & \textcolor{teal}{+4} & \textcolor{teal}{+6.6} \\
        \bottomrule
    \end{tabular}}
\end{table}

\vspace{-5pt}

\section{Conclusion}
\label{sec:conclusion}
\vspace{-5pt}
In this paper, we presented GeoSem-WAM, a geometry- and semantics-enhanced world-action model designed as a plug-and-play module for robot manipulation. By attaching auxiliary DPT-style prediction heads to intermediate video-expert tokens during training, GeoSem-WAM learns from dense geometric and semantic supervision while keeping the action inference pipeline efficient and unchanged at deployment. This structured, training-only supervision allows the model to capture not only visual dynamics, but also task-relevant spatial layouts and object-level semantics, enhancing spatial motion awareness and task logical reasoning without adding test-time computational overhead. 
While effective, GeoSem-WAM has two primary limitations. First, the auxiliary DPT head currently relies on pixel-level annotations, which are scarce in real RGB-only datasets and often require generating potentially noisy pseudo-labels. Future work will explore leveraging self-supervised features from foundation models (e.g., DINO \cite{simeoni2025dinov3}) to implicitly extract spatial and categorical priors directly from raw RGB videos. Second, jointly optimizing heterogeneous loss functions increases training complexity and poses a risk of gradient conflicts. To address this, we aim to integrate gradient deconfliction algorithms, such as gradient surgery \cite{yu2020gradient}, to mitigate task interference. Overall, our findings suggest that structured geometric and semantic prediction serves as a valuable auxiliary signal for learning richer representations in world-action modeling.


\clearpage
\acknowledgments{If a paper is accepted, the final camera-ready version will (and probably should) include acknowledgments. All acknowledgments go at the end of the paper, including thanks to reviewers who gave useful comments, to colleagues who contributed to the ideas, and to funding agencies and corporate sponsors that provided financial support.}


\bibliography{example}  

\appendix

\section{Simulation Environments and Multi-Modal Observations}
\begin{figure}[t]
    \centering
    \includegraphics[width=1.0\linewidth]{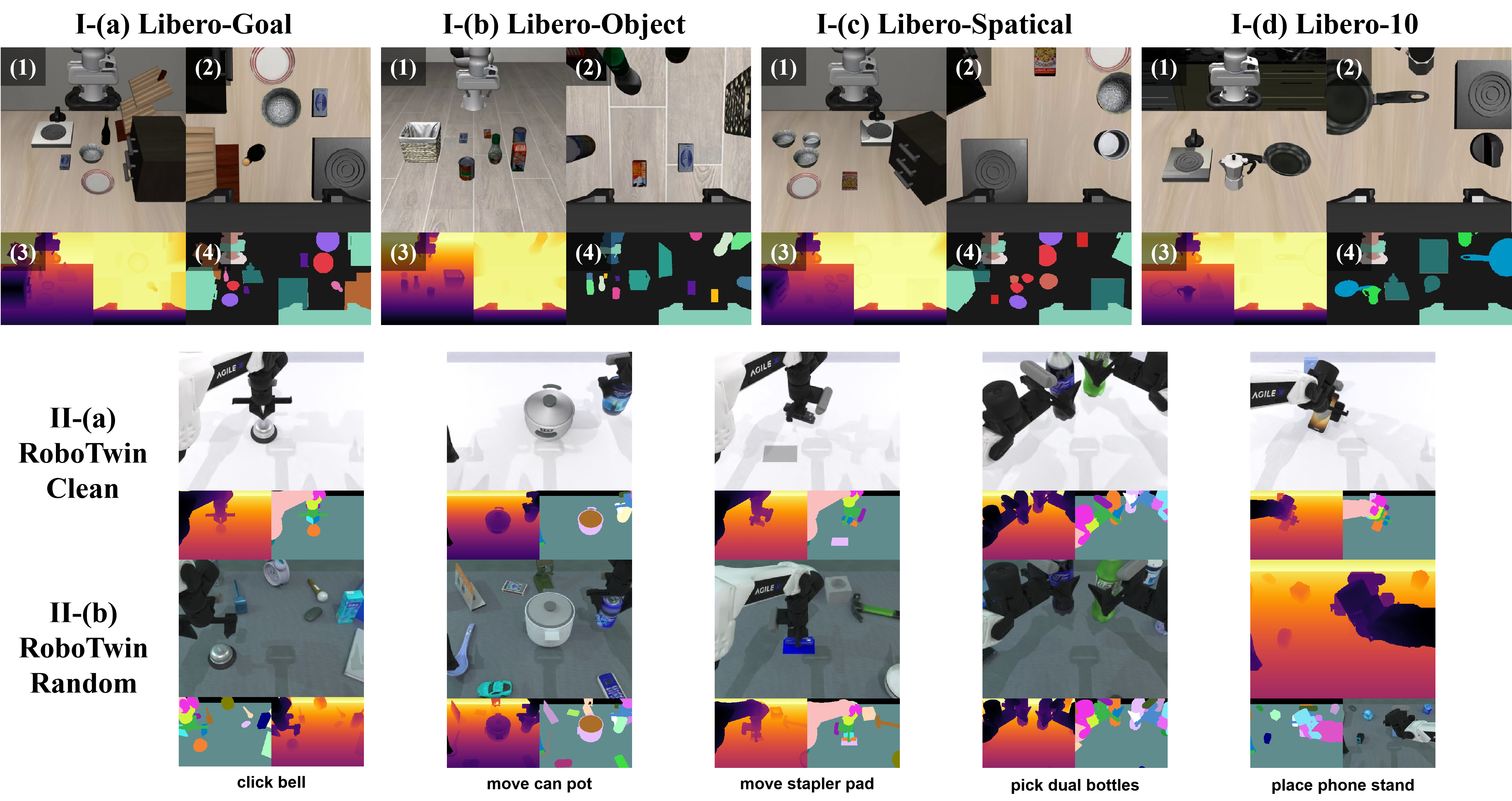}
    \caption{\textbf{Overview of simulation environments and multi-modal observations.}
    \textbf{(I) Libero benchmark tasks:} (a) Libero-Goal, (b) Libero-Object, (c) Libero-Spatial, and (d) Libero-10. Each task is visualized with RGB observations (1,2), paired with corresponding depth maps (3) and semantic segmentation masks (4).
    \textbf{(II) RoboTwin benchmark tasks:} (a) Clean environment, and (b) Random environment. Representative tasks include click bell, move can pot, move stapler pad, pick dual bottles, and place phone stand, with paired RGB, depth, and semantic observations.}
    \label{fig:sim_libero}
\end{figure}

We evaluate our GeoSem-WAM on two challenging simulation benchmarks, Libero~\cite{liu2023libero} and RoboTwin~\cite{chen2025robotwin}, as illustrated in Figure~\ref{fig:sim_libero}.

\paragraph{Libero Benchmark.} Libero is a household manipulation benchmark with four task suites of increasing complexity:
\begin{itemize}
    \item \textit{Libero-Goal}: Tasks with varying goal specifications, requiring the policy to adapt to different target states.
    \item \textit{Libero-Object}: Tasks with diverse object types and configurations, testing object-centric manipulation capabilities.
    \item \textit{Libero-Spatial}: Tasks requiring fine-grained spatial reasoning and relative positioning of objects.
    \item \textit{Libero-10}: A combined suite of 10 long-horizon household tasks, representing the most challenging setting.
\end{itemize}
For each task, we collect multi-modal observations including third-person RGB images, depth maps, and pixel-level semantic segmentation masks, as shown in Figure~\ref{fig:sim_libero}-I. These modalities provide complementary geometric and semantic cues for policy learning.

\begin{table*}[h]
\centering
\caption{Quantitative results for each task on the RoboTwin 2.0 simulation benchmark, covering 50 bimanual manipulation tasks with two difficulty levels.}
\resizebox{\textwidth}{!}{%
\begin{tabular}{lcccccccccccccc}
\toprule
\multirow{2}{*}{Model} 
& \multicolumn{2}{c}{GeoSem-WAM (Ours)} 
& \multicolumn{2}{c}{Fast-WAM} 
& \multicolumn{2}{c}{LingBot} 
& \multicolumn{2}{c}{Pi\_05} 
& \multicolumn{2}{c}{Pi\_0} 
& \multicolumn{2}{c}{X-VLA} 
& \multicolumn{2}{c}{Motus} \\
\cmidrule(lr){2-3} \cmidrule(lr){4-5} \cmidrule(lr){6-7} 
\cmidrule(lr){8-9} \cmidrule(lr){10-11} \cmidrule(lr){12-13} \cmidrule(lr){14-15}
Task Type 
& clean & random & clean & random & clean & random 
& clean & random & clean & random & clean & random & clean & random \\
\midrule
Adjust Bottle & 100 & 100 & 100 & 100 & 90 & 94 & 100 & 99 & 99 & 95 & 100 & 99 & 89 & 93 \\
Beat Block Hammer & 98 & 98 & 99 & 97 & 96 & 98 & 96 & 93 & 79 & 84 & 92 & 88 & 95 & 88 \\
Blocks Ranking RGB & 100 & 98 & 100 & 100 & 99 & 98 & 92 & 85 & 80 & 63 & 83 & 83 & 99 & 97 \\
Blocks Ranking Size & 91 & 96 & 94 & 98 & 94 & 96 & 49 & 26 & 14 & 5 & 67 & 74 & 75 & 63 \\
Click Alarmclock & 100 & 100 & 100 & 100 & 99 & 100 & 98 & 89 & 77 & 68 & 99 & 99 & 100 & 100 \\
Click Bell & 100 & 100 & 100 & 100 & 100 & 100 & 99 & 66 & 71 & 48 & 100 & 100 & 100 & 100 \\
Dump Bin Big Binbin & 96 & 95 & 97 & 96 & 89 & 96 & 92 & 97 & 88 & 83 & 79 & 77 & 95 & 91 \\
Grab Roller & 100 & 100 & 100 & 100 & 100 & 100 & 100 & 100 & 98 & 94 & 100 & 100 & 100 & 100 \\
Handover Block & 96 & 81 & 95 & 81 & 99 & 78 & 66 & 57 & 47 & 31 & 73 & 37 & 86 & 73 \\
Handover Mic & 100 & 100 & 99 & 100 & 94 & 96 & 98 & 97 & 97 & 97 & 0 & 0 & 78 & 63 \\
Hanging Mug & 72 & 67 & 58 & 62 & 40 & 28 & 18 & 17 & 14 & 11 & 23 & 27 & 38 & 38 \\
Lift Pot & 100 & 100 & 100 & 100 & 100 & 99 & 96 & 85 & 80 & 72 & 99 & 100 & 96 & 99 \\
Move Can Pot & 90 & 95 & 90 & 88 & 94 & 97 & 51 & 55 & 68 & 48 & 89 & 86 & 34 & 74 \\
Move Pillowbottle Pad & 99 & 98 & 100 & 99 & 99 & 99 & 84 & 61 & 67 & 46 & 73 & 71 & 93 & 96 \\
Move Playingcard Away & 100 & 100 & 100 & 100 & 100 & 99 & 96 & 84 & 74 & 65 & 93 & 98 & 100 & 96 \\
Move Stapler Pad & 73 & 63 & 77 & 64 & 91 & 79 & 56 & 42 & 41 & 24 & 78 & 73 & 83 & 85 \\
Open Laptop & 99 & 100 & 98 & 100 & 92 & 94 & 90 & 96 & 71 & 81 & 93 & 100 & 95 & 91 \\
Open Microwave & 75 & 50 & 62 & 45 & 82 & 86 & 34 & 77 & 4 & 32 & 79 & 71 & 95 & 91 \\
Pick Diverse Bottles & 87 & 86 & 80 & 85 & 89 & 82 & 81 & 71 & 69 & 31 & 58 & 36 & 90 & 91 \\
Pick Dual Bottles & 100 & 97 & 100 & 96 & 100 & 99 & 93 & 63 & 59 & 37 & 47 & 36 & 96 & 90 \\
Place A2B Left & 93 & 95 & 95 & 93 & 97 & 93 & 87 & 82 & 43 & 47 & 48 & 49 & 82 & 79 \\
Place A2B Right & 95 & 95 & 93 & 99 & 97 & 95 & 87 & 84 & 39 & 34 & 36 & 36 & 90 & 87 \\
Place Bread Basket & 90 & 92 & 91 & 93 & 97 & 95 & 77 & 64 & 62 & 46 & 81 & 71 & 91 & 94 \\
Place Bread Skillet & 91 & 98 & 90 & 93 & 95 & 90 & 85 & 66 & 66 & 49 & 77 & 67 & 86 & 83 \\
Place Burger Fries & 96 & 100 & 96 & 99 & 97 & 95 & 94 & 87 & 81 & 76 & 94 & 94 & 98 & 98 \\
Place Can Basket & 70 & 70 & 71 & 69 & 81 & 84 & 62 & 62 & 55 & 46 & 49 & 52 & 81 & 76 \\
Place Cans Plasticbox & 99 & 99 & 99 & 96 & 100 & 99 & 94 & 84 & 63 & 45 & 97 & 98 & 98 & 94 \\
Place Container Plate & 98 & 98 & 96 & 100 & 99 & 97 & 99 & 95 & 97 & 92 & 97 & 95 & 98 & 99 \\
Place Dual Shoes & 93 & 91 & 94 & 88 & 94 & 89 & 75 & 75 & 59 & 51 & 79 & 88 & 93 & 87 \\
Place Empty Cup & 100 & 100 & 100 & 100 & 100 & 100 & 100 & 99 & 91 & 85 & 100 & 98 & 99 & 98 \\
Place Fan & 97 & 96 & 96 & 96 & 99 & 93 & 87 & 85 & 66 & 71 & 80 & 75 & 91 & 87 \\
Place Mouse Pad & 89 & 89 & 83 & 89 & 93 & 96 & 60 & 39 & 20 & 20 & 70 & 70 & 66 & 68 \\
Place Object Basket & 89 & 85 & 89 & 88 & 91 & 88 & 80 & 76 & 67 & 70 & 44 & 39 & 81 & 87 \\
Place Object Scale & 92 & 92 & 90 & 97 & 96 & 95 & 86 & 80 & 57 & 52 & 52 & 74 & 88 & 85 \\
Place Object Stand & 91 & 91 & 90 & 94 & 99 & 96 & 91 & 85 & 82 & 68 & 86 & 88 & 98 & 97 \\
Place Phone Stand & 98 & 99 & 97 & 99 & 97 & 97 & 81 & 81 & 49 & 53 & 88 & 87 & 87 & 86 \\
Place Shoe & 97 & 99 & 96 & 99 & 98 & 98 & 92 & 93 & 76 & 76 & 96 & 95 & 99 & 97 \\
Press Stapler & 94 & 96 & 90 & 97 & 85 & 82 & 87 & 83 & 44 & 37 & 92 & 98 & 93 & 98 \\
Put Bottles Dustbin & 92 & 88 & 95 & 90 & 87 & 91 & 84 & 79 & 65 & 56 & 74 & 77 & 81 & 79 \\
Put Object Cabinet & 92 & 87 & 94 & 89 & 85 & 87 & 80 & 79 & 73 & 60 & 46 & 48 & 88 & 71 \\
Rotate QRcode & 96 & 95 & 93 & 89 & 96 & 91 & 89 & 87 & 74 & 70 & 34 & 33 & 89 & 73 \\
Scan Object & 89 & 89 & 89 & 92 & 96 & 91 & 72 & 65 & 55 & 42 & 14 & 36 & 67 & 66 \\
Shake Bottle Horizontally & 100 & 100 & 100 & 100 & 100 & 99 & 99 & 99 & 98 & 92 & 100 & 100 & 100 & 98 \\
Shake Bottle & 100 & 100 & 100 & 100 & 100 & 97 & 99 & 97 & 94 & 91 & 99 & 100 & 100 & 97 \\
Stack Blocks Three & 96 & 98 & 95 & 97 & 99 & 98 & 91 & 76 & 72 & 52 & 6 & 10 & 91 & 95 \\
Stack Blocks Two & 100 & 100 & 100 & 100 & 100 & 98 & 97 & 100 & 93 & 79 & 92 & 87 & 100 & 98 \\
Stack Bowls Three & 91 & 83 & 80 & 81 & 86 & 83 & 77 & 71 & 77 & 75 & 76 & 86 & 79 & 87 \\
Stack Bowls Two & 93 & 98 & 92 & 98 & 94 & 98 & 95 & 96 & 94 & 95 & 96 & 93 & 98 & 98 \\
Stamp Seal & 92 & 95 & 90 & 94 & 96 & 97 & 79 & 55 & 46 & 33 & 76 & 82 & 93 & 92 \\
Turn Switch & 58 & 65 & 61 & 59 & 44 & 45 & 62 & 54 & 41 & 42 & 40 & 61 & 84 & 78 \\
\midrule
Average & \textbf{92.94} & \textbf{92.14} & 91.88 & 91.78 & 92.9 & 91.5 & 82.74 & 76.76 & 65.92 & 58.4 & 72.88 & 72.84 & 88.52 & 87.02 \\
\bottomrule
\end{tabular}
}
\label{tab:robo_twin_sim}
\end{table*}

\paragraph{RoboTwin Benchmark.} 
RoboTwin is a large-scale simulation data generation and benchmarking platform for bimanual robotic manipulation, designed to address the challenges of scarce high-quality training data and difficult sim-to-real transfer. The platform integrates automated expert demonstration generation, large-scale multi-modal datasets, and standardized evaluation systems. Its core features include: a 3D object library containing 731 fine-grained object instances across 147 categories, a closed-loop expert code synthesis pipeline based on multimodal large language models, structured domain randomization across five dimensions (clutter, lighting, background texture, tabletop height, and language instructions), and a standardized benchmark covering 50 bimanual tasks with support for 5 robot embodiments, along with an open dataset of over 100, 000 expert trajectories and clean/random evaluation protocols.
We evaluate on two settings:
\begin{itemize}
    \item \textit{Clean Environment}: A controlled setting with minimal visual clutter, serving as a baseline for task performance.
    \item \textit{Random Environment}: A highly cluttered setting with randomly placed distractors, evaluating the policy's robustness to visual noise and background distractions.
\end{itemize}
The evaluation details are shown in Table \ref{tab:robo_twin_sim}, which demonstrates that our GeoSem-WAM achieves the best overall success rate compared to previous SOTA methods.
Representative tasks include click bell, move can pot, move stapler pad, pick dual bottles, and place phone stand. As shown in Figure~\ref{fig:sim_libero}-II, each task provides synchronized RGB, depth, and semantic observations to support multi-modal policy learning.

\section{Example Episodes of Real World Experiments on Franka}
\begin{figure}[t]
    \centering
    \includegraphics[width=1.0\linewidth]{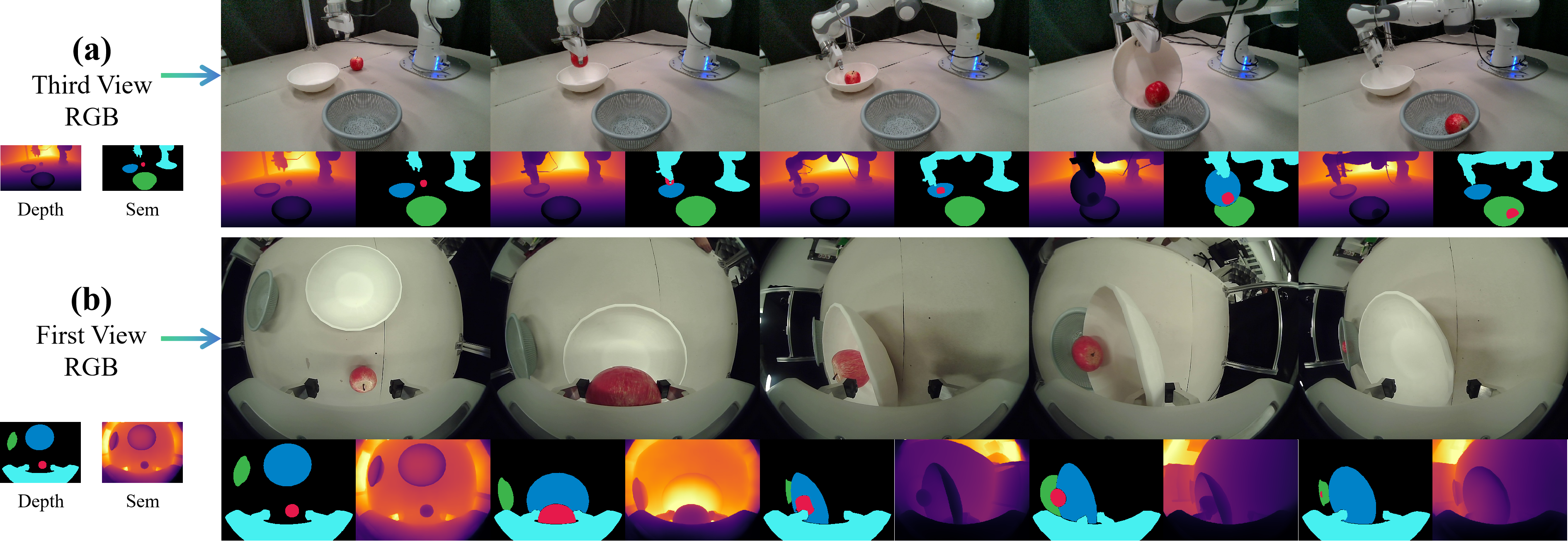}
    \caption{\textbf{Example Episodes of Real World Experiments on Franka.} 
    \textbf{Step-by-step demonstration of the Pick-Pour task (I-d):} multi-modal observations from both third-person (a) and first-person (ego-centric) (b) views, including RGB, geometry, and semantic segmentation at each key stage of the pick-place-pour sequence.}
    \label{fig:real_franka_pour}
\end{figure}

Figure~\ref{fig:real_franka_pour} presents the detailed execution flow of the Pick-Pour task. It displays synchronized third-person and ego-centric first-person observations, along with corresponding geometry and semantic segmentation outputs throughout the whole manipulation process.

\end{document}